\PassOptionsToPackage{table,dvipsnames}{xcolor}

\documentclass[acmsmall]{acmart}

\AtBeginDocument{%
  }

\setcopyright{acmlicensed}
\copyrightyear{2018}
\acmYear{2018}
\acmDOI{XXXXXXX.XXXXXXX}

\acmJournal{JACM}
\acmVolume{37}
\acmNumber{4}
\acmArticle{111}
\acmMonth{8}

\usepackage{amsthm}
\usepackage{amsmath}
\usepackage{bm}
\usepackage{multirow}
\usepackage{enumitem}
\usepackage{hyperref}       
\hypersetup{
    colorlinks=true,
    linkcolor=blue,
    urlcolor=blue,
    citecolor=purple,
}
\usepackage{url}            
\usepackage{booktabs}       
\usepackage{amssymb}
\usepackage{makecell}
\usepackage{color}
\usepackage{mathrsfs}
\usepackage{pifont}
\usepackage{xcolor}
\definecolor{newcolor}{rgb}{.8,.349,.1}
\usepackage{hyperref}

\definecolor{gray}{rgb}{0.35,0.35,0.35}
\definecolor{pinklight}{rgb}{1,0,1}
\definecolor{yellowlight}{rgb}{1,1,0}
\definecolor{blue}{rgb}{0,0,1}
\definecolor{red}{rgb}{1,0,0}
\definecolor{orange}{rgb}{0.75, 0.4, 0}
\definecolor{purple}{rgb}{0.5, 0.0, 0.5}
\definecolor{black}{rgb}{0,0,0}

\newcommand{\jm}[1]{{\color{black}#1}\normalfont}

\usepackage{wrapfig}

\usepackage{setspace}
\usepackage{color}

\begin{document}

\title{ShapeMoiré: Channel-Wise Shape-Guided Network for Image Demoiréing}


\author{Jinming Cao}
\affiliation{%
  \institution{National University of Singapore}
  \country{Singapore}}
\email{jinming.ccao@gmail.com}

\author{Sicheng Shen}
\affiliation{%
  \institution{National University of Singapore}
  \country{Singapore}}
\email{sshen@u.nus.edu}

\author{Qiu Zhou}
\affiliation{%
  \institution{Beijing university of posts and telecommunications}
  \city{Beijing}
  \country{China}}
\email{zhouqiulv@gmail.com}

\author{Yifang Yin}
\affiliation{%
  \institution{Institute for Infocomm Research (I$^2$R), A*STAR}
  \country{Singapore}
}
\email{yin\_yifang@i2r.a-star.edu.sg}

\author{Yangyan Li}
\affiliation{%
 \institution{Alibaba Group}
 \country{China}}
\email{yangyan.lee@gmail.com}

\author{Roger Zimmermann}
\affiliation{%
  \institution{National University of Singapore}
  \country{Singapore}}
  \email{rogerz@comp.nus.edu.sg}



\renewcommand{\shortauthors}{Trovato et al.}

\begin{abstract}
Photographing optoelectronic displays often introduces unwanted moiré patterns due to analog signal interference between the pixel grids of the display and the camera sensor arrays. 
This work identifies two problems that are largely ignored by existing image demoiréing approaches:
1) moiré patterns vary across different channels (RGB); 2) repetitive patterns are constantly observed.
However, employing conventional convolutional (CNN) layers cannot address these problems.
Instead, this paper presents the use of our recently proposed \emph{Shape} concept.
It was originally employed to model consistent features from fragmented regions, particularly when identical or similar objects coexist in an RGB-D image. 
Interestingly, we find that the \emph{Shape} information effectively captures the moiré patterns in artifact images.
Motivated by this discovery, we propose a \jm{new method,} ShapeMoiré, \jm{for} image demoiréing.
Beyond modeling shape features at the patch-level, we further extend this to the global image-level and design a novel Shape-Architecture.
Consequently, our proposed method, equipped with both ShapeConv and Shape-Architecture, can be seamlessly integrated into existing approaches without introducing \jm{any} additional parameters or computation overhead during inference.
We conduct extensive experiments on four widely used datasets, and the results demonstrate that our ShapeMoiré achieves state-of-the-art performance, particularly in terms of the PSNR metric.
We then apply our method across four popular architectures to showcase its generalization capabilities.
\jm{Moreover, to further validate its generality beyond the demoiréing task, we apply ShapeMoiré to the image deblurring task, where it continues to deliver consistent performance gains.
Finally, experiments on real-world images captured by smartphones confirm the robustness and practical applicability of ShapeMoiré in challenging demoiréing scenarios.}
We open-sourced an implementation of ShapeMoiré in PyTorch at \href{https://github.com/SichengS/ShapeMoire}{\underline{https://github.com/ShapeMoire}}.
\end{abstract}



\received{20 February 2007}
\received[revised]{12 March 2009}
\received[accepted]{5 June 2009}

\maketitle

\section{Introduction}
\label{sec:intro}

\begin{figure}[t!]
	\centering
	\centering
	\includegraphics[width=0.88\textwidth]{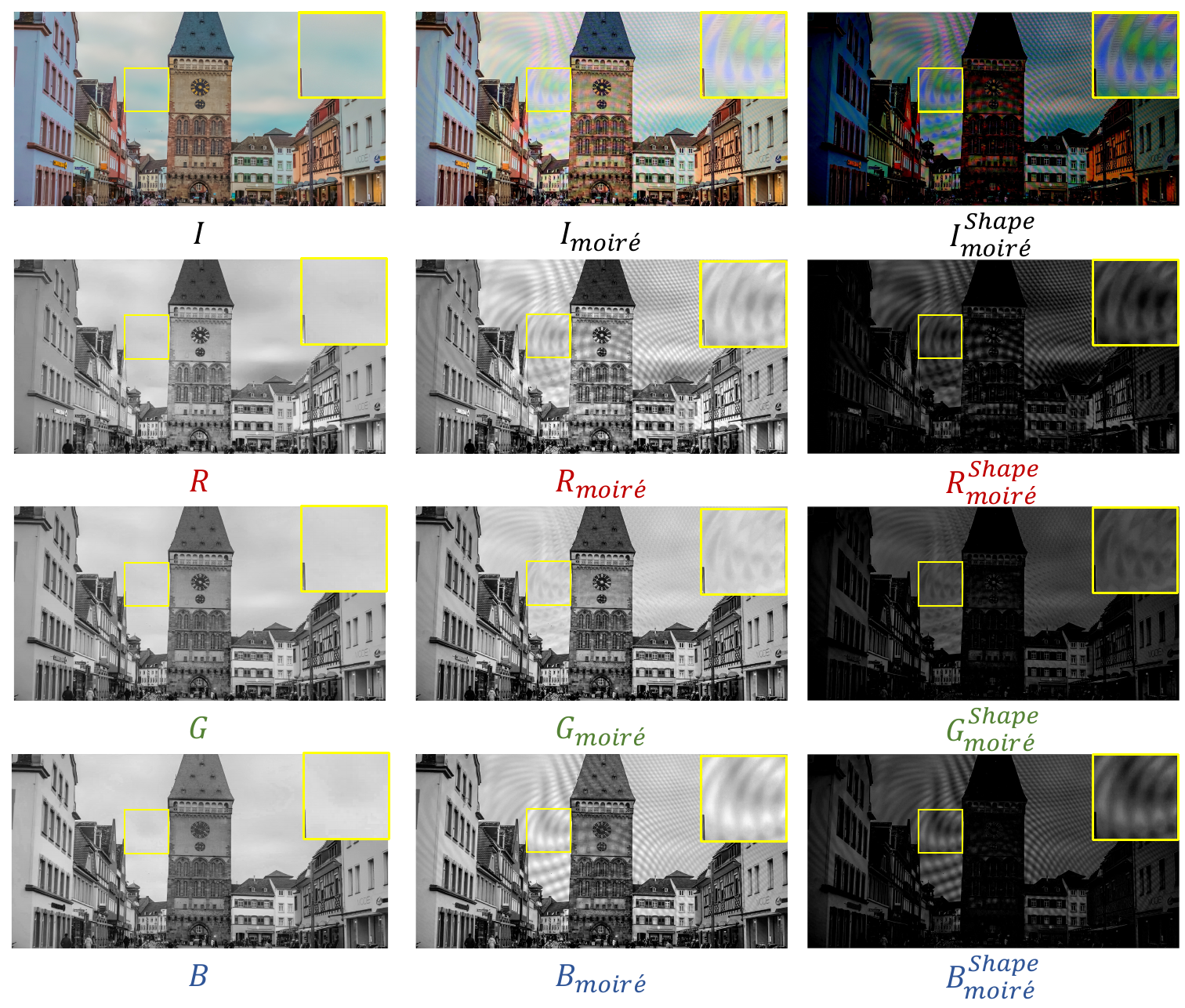}
 
	\caption{Visual demonstration of the moiré pattern in the given image and its separate manifestations in the individual RGB channels. 
Left column: the original image $I$, where the color in each channel is uniformly and smoothly distributed; Middle column: it can be observed that the three channels of $I_{moir\acute{e}}$ display distinct moiré patterns; Right column: the shapes of moiré patterns in the \emph{shape} space ($R^{Shape}_{moir\acute{e}}$, $G^{Shape}_{moir\acute{e}}$ and $B^{Shape}_{moir\acute{e}}$) remain consistent with the patterns in the different channels of the original moiré image ($R_{moir\acute{e}}$, $G_{moir\acute{e}}$ and $B_{moir\acute{e}}$), but with enhanced structural integrity.
}
	\label{fig:teaser}
 \end{figure}

Moiré artifacts manifest as wavy or rippled distortions in digital photographs, arising from the interference between a camera’s color filter array and the sub-pixel structure of electronic displays. 
These unwanted patterns significantly degrade visual quality, posing challenges in multimedia applications such as mobile photography~\cite{rawat2015context} and cross-device content reproduction~\cite{ng2017smart}.
Fig.~\ref{fig:teaser} illustrates a digital photo with such moiré patterns ($I_{moir\acute{e}}$), which make the original image ($I$) appear opaque and distorted when displayed on a screen. 
Notably, the irregular shape and \jm{broad frequency ranges} of moiré patterns distinguish demoiréing from other image restoration tasks such as super-resolution~\cite{zhang2015image,wei2023taylor} and reducing JPEG compression artifacts~\cite{dong2015compression}.

Moiré pattern removal has attracted extensive research interests over the past few years~\cite{yang2017textured,he2019mop,he2020fhde,yu2022towards,yue2022recaptured}. 
Initially, efforts were devoted to collecting synthetic data~\cite{yuan2019aim} or genuine images with low resolution~\cite{sun2018moire}.
Subsequent datasets expanded the image resolution to 1080P~\cite{he2020fhde} or even ultra-high-definition (4K)~\cite{yu2022towards}, posing greater challenges in removing moirés without compromising the pristine features of the images.
To address this, prior methods have mostly relied on rich side information, such as complementary visual attributes~\cite{he2019mop} and aliasing frequencies~\cite{liu2020wavelet,zheng2020image}.
Despite the advancements in existing methods, they often treat the input image from a holistic perspective, especially without discriminating between different channels (i.e., RGB channels).
However, as observed in Fig.~\ref{fig:teaser}, unlike the uniform and smooth color distribution in each channel ($R$, $G$ and $B$) of the original image $I$, these three input channels ($R_{moir\acute{e}}$, $G_{moir\acute{e}}$ and $B_{moir\acute{e}}$) typically exhibit distinct moiré patterns.
As a result, processing them equally with convolutional layers confuses the model's focus on demoiréing, resulting in sub-optimal outcomes.
Furthermore, existing literature often overlooks the repetitive patterns that are consistently observed.

Our idea in this work is inspired by a recent prevailing concept—\emph{Shape}—in RGB-D segmentation~\cite{cao2021shapeconv}.
Unlike traditional CNNs, ShapeConv~\cite{cao2021shapeconv} can distinguish between different channels and is adept at modeling repetitive shapes, such as two identical chairs in a depth image from RGB-D segmentation.
In the context of moiré patterns in an image, as demonstrated in Fig.~\ref{fig:teaser}, the moiré image ($I_{moir\acute{e}}$) exhibits frequent similar or repetitive patterns, unlike the uniform color distribution in the original image ($I$). This type of information is precisely what \emph{shape} excels at expressing. 
Additionally, the features of moiré patterns in the \emph{shape} space ($R^{Shape}_{moir\acute{e}}$, $G^{Shape}_{moir\acute{e}}$ and $B^{Shape}_{moir\acute{e}}$) remain consistent across different channels of the original moiré image ($R_{moir\acute{e}}$, $G_{moir\acute{e}}$ and $B_{moir\acute{e}}$), but \jm{exhibit} enhanced structural integrity. 
This \jm{observation} motivates us to approach the image demoiré problem from the \emph{shape} perspective.

Therefore, we propose a novel method named ShapeMoiré to extend our ShapeConv for image demoiring. 
Our ShapeMoiré operates on both \jm{the} patch\footnote{The operational unit of input features for the convolutional layer, whose spatial size matches that of the convolution kernel.}-level and image-level shapes.
This results in two key components of our proposed ShapeMoiré: the Shape-aware convolutional layer (ShapeConv) and \jm{the} Shape-aware network architecture (Shape-Architecture).
Specifically, at the patch-level, we first decompose a patch into two separate components, i.e., a \emph{base-component} and a \emph{shape-component}. 
We then employ two operations, namely, base-product and shape-product, to process these components \jm{separately}, with two learnable weights: the base-kernel and the shape-kernel. The output\jm{s} from these operations are then combined in an \jm{additive} manner to form a shape-aware patch, which is subsequently convolved with a standard convolutional kernel. 
Unlike the original patch, the shape-aware patch is capable of adaptively learning the shape \jm{characteristics} with the shape-kernel, while the base-kernel serves to balance the contributions of the \emph{shape} and the \emph{base} components in the final feature representation.
\jm{At} the extended image-level, we incorporate an additional training stream to enhance the model's focus on shape information during training. This is achieved by adding a loss function for the shape branch, thereby updating the parameters with an emphasis on shape information.
In a nutshell, the former replaces the conventional convolutional layer with a delicately designed ShapeConv that involves the modeling of shape features, and the latter employs the shape information of raw pixels to further enhance the baseline architecture.
It is worth noting that our ShapeMoiré can be seamlessly integrated into most baseline architectures that are equipped with convolutional layers as building blocks.
In addition, benefiting from its original virtue, our ShapeMoiré results in no incremental parameters and \jm{only} negligible computational cost.

To evaluate the effectiveness of our proposed method, we conducted extensive experiments on four well-established datasets: LCDMoiré~\cite{yuan2019aim}, TIP2018~\cite{sun2018moire}, FHDMi~\cite{he2020fhde}, and UHDM~\cite{yu2022towards}. 
Our experimental results demonstrate that ShapeMoiré \jm{consistently} outperforms existing state-of-the-art \jm{methods} by a significant margin.
To assess its generalization capabilities, we applied ShapeMoiré to various widely adopted demoiréing architectures~\cite{sun2018moire,liu2020wavelet,yu2022towards} and observed improved results compared to baselines.
\jm{Furthermore, to validate the method’s broader applicability beyond demoiréing, we extended our experiments to an additional image restoration task—image deblurring. Results on the DPDD dataset~\cite{abuolaim2020defocus} confirm that ShapeMoiré remains effective in this new context, indicating good cross-task generalization.}
Notably, qualitative results highlight that ShapeMoiré remarkably enhances moiré pattern removal, even when applied to real-world images captured using a smartphone. This underscores both the effectiveness on public datasets and the practical applicability of our ShapeMoiré in real-world scenarios.

The work on ShapeConv has previously been published at  ICCV~\cite{cao2021shapeconv}. This extended manuscript expands upon the original version in several key aspects:
\begin{itemize}
\item  We broaden the application of the notion of \emph{Shape} beyond its original use in RGB-D segmentation. 
\jm{
To the best of our knowledge, this is the first time the concept has been effectively applied to both the moiré pattern removal and image deblurring tasks in 2D image restoration.}

\item  Beyond employing patch-level ShapeConv, we introduce an image-level Shape-Architecture for image demoiréing. 
\jm{
Through comprehensive ablation studies and cross-task experiments, we demonstrate that integrating \emph{Shape} information at multiple levels consistently enhances performance on diverse image restoration tasks.}

\item  The proposed ShapeMoiré method, which incorporates both ShapeConv and Shape-Architecture, represents a versatile approach. We conduct experiments using various existing methods as baselines, all of which effectively enhance baseline performance. Notably, both ShapeConv and Shape-Architecture introduce ZERO additional parameters and \jm{incur only} negligible computational overhead during inference.

\end{itemize}
\section{Related Work}
\label{sec:rw}

\subsection{Image Demoiréing}
Removing moiré patterns constitutes a form of image denoising aimed at mitigating the destructive artifacts arising from frequency interference during image capture. 
Unlike other types of noise, such as flickering\footnote{\jm{\url{https://digitalanarchy.com/Flicker/main.html}}} induced by photography, there is currently no \jm{standardized} tool for effectively eliminating moiré patterns.
Recently, numerous benchmark datasets and approaches have been introduced to address the challenge of image demoiréing.

\paragraph{Datasets}
Liu et al.~\cite{liu2018demoir} initiated the task by curating a synthetic dataset that simulates the process of photographing with a camera. Building upon this approach, Yuan et al.~\cite{yuan2019aim} developed a large-scale synthetic dataset, namely, LCDMoiré, for further image demoiréing~\cite{yuan2019aim,zheng2020image,cheng2019multi}. 
However, the transition from synthetic training to real-world scenarios presents a significant obstacle due to the inherent gap between simulation and reality. 
Sun et al.~\cite{sun2018moire} addressed this challenge by introducing the TIP2018 dataset, which utilizes real-world moiré images. Following this approach, Yue et al.\cite{yue2022recaptured} compiled a well-aligned dataset of raw moiré image.
Nevertheless, these images often have low resolutions, making it difficult for the accompanying methods to adapt to high-resolution images. 
To tackle this issue, He et al.~\cite{he2020fhde} curated the FHDMi dataset with a resolution of 1080P. However, as the demand for image quality continues to increase, the practicality of 1080p resolution may diminish, especially when considering the ultra-high-definition (4K) images captured by modern mobile cameras. Recognizing these challenges and limitations, Yu et al.~\cite{yu2022towards} collected the first ultra-high-definition demoiréing dataset (UHDM), which contains 5,000 real-world 4K resolution image.

\paragraph{Methods} Pertaining to moiré pattern removal methods, current approaches often endeavor to design specific architectures for this task.
For example, He et al.~\cite{he2019mop} proposed to classify moiré patterns using manually annotated category labels. 
Some more approaches~\cite{zheng2020image,liu2020wavelet} have focused on the frequency domain via de-aliasing.
In order to remove moiré without compromising the clarity of underlying image details, Sun et al.~\cite{sun2018moire} implemented a multi-scale network, DMCNN, effectively addressing moiré patterns in practical scenarios. Building upon DMCNN, He et al.~\cite{he2020fhde} developed the multi-stage framework FHD$e^2$Net to handle a wider pattern scale range and preserve fine details. 
For 4K images, ESDNet~\cite{yu2022towards} implements an efficient module that is semantic-aligned and scale-aware to address the scale variation of moiré patterns.

Unlike previous methods, which treat the entire image as a whole without distinguishing between different channels, Yang et al.~\cite{yang2017textured} also observed differences in moiré patterns across different RGB channels. Specifically, they found that the red (R) and blue (B) channels are more heavily affected by moiré artifacts compared to the green (G) channel. Therefore, they proposed to remove moiré artifacts in the R and B channels using guided filtering based on the texture layer obtained from the G channel. In contrast to the method in~\cite{yang2017textured}, our approach does not involve special treatment of a specific channel but instead presents a universal method that discriminates between different channels. Moreover,~\cite{yang2017textured} is a specialized design, while our proposed ShapeMoiré stands out as a versatile approach. It can seamlessly integrate with existing methods, enhancing model performance with ZERO increase of memory and computation.

\subsection{Image Restoration}
Image restoration aims to reconstruct the original, clean and high-quality image from degraded versions, serving as a core task in computer vision and multimedia computing. 
This task encompasses various processes, including image super-resolution (SR)~\cite{jiang2016noise}, image denoising~\cite{chen2015advanced,yan2020depth,xu2023cur}, reducing JPEG compression artifacts and so on.
A plethora of learning-based image restoration models~\cite{liang2021swinir,zamir2021multi,zamir2021multi}, have gained popularity due to their impressive performance.
These models typically leverage large-scale paired datasets to learn mappings between low-quality and high-quality images, employing encoder-decoder structures~\cite{ronneberger2015u} or hierarchical architectures~\cite{sun2018moire,zhang2019deep}. 
Building upon pioneering works such ase SRCNN~\cite{dong2014learning} (for SR), DnCNN~\cite{zhang2017beyond} (for denoising), and ARCNN~\cite{dong2015compression} (for reducing JPEG artifacts), numerous CNN-based models~\cite{lai2017deep,zhang2018learning,wang2019learning,guo2023modality} have emerged. 
These models enhance representational capacity by incorporating larger and deeper neural architectures, such as residual blocks~\cite{he2016deep}, dense blocks~\cite{huang2017densely}, and others~\cite{lai2017deep}. \jm{In recent years, transformer-based and attention-driven frameworks have further advanced the field.
Restormer~\cite{zamir2022restormer} proposes a multi-Dconv head transposed attention mechanism tailored for image restoration, achieving state-of-the-art results on multiple benchmarks.
Uformer~\cite{wang2022uformer} unifies transformer and U-Net architectures to improve global context modeling while maintaining efficiency.
Guo et al.~\cite{guo2024mambair} introduces a simple but effective baseline, named MambaIR, which introduces both local enhancement and channel attention, yet surprisingly achieves competitive performance on several restoration tasks.
Lin et al.~\cite{lin2024improving} maps the degraded images into textual representations for removing the degradations, and then convert the restored textual representations into a guidance image for assisting image restoration.

However, while these methods target general image restoration tasks—such as image super-resolution, image denoising, and JPEG compression artifact reduction—the denoising types they address are mostly spatial-domain noises like Gaussian color image denoising~\cite{zamir2022restormer,guo2024mambair}, real image denoising~\cite{guo2024mambair}, deraining~\cite{wang2022uformer}, and deblurring~\cite{lin2024improving}.
Although moiré patterns are also a type of noise, they are frequency-domain artifacts with unique structural characteristics and thus have not been effectively tackled by these existing methods.}

\subsection{Shape Concept and Its Applications}

The concept of \emph{Shape} was initially introduced in the domain of RGB-D segmentation in a previously published paper~\cite{cao2021shapeconv}, where it is defined as the disparity between local depth values and the overall distance of the region from the observation point.
One key observation from this paper is that, in spaces where identical or similar objects coexist, \emph{Shape} can extract more consistent features compared to raw pixel values. 
To leverage this insight, we designed ShapeConv to replace traditional convolution layers, allowing for the processing of data from various modalities while emphasizing \emph{Shape} attributes. 
Subsequent studies ~\cite{xiong2023gamus,zhong2023combining} extended the application to remote sensing data, showcasing \emph{Shape}'s superiority in feature extraction across RGB+nDSM (primarily representing height features) data modalities. Specifically, Xiong et al.~\cite{xiong2023gamus} utilized ShapeConv for semantic segmentation tasks on the their collected GAMUS dataset, which features numerous repetitive buildings. 
Zhong et al.~\cite{zhong2023combining} created an ocean coral dataset characterized by repetitive texture patterns within the corals. By integrating ShapeConv as a fundamental component of their model, they achieved superior performance.
Beyond the RGB+X data, \cite{zhou2024sogdet} employs ShapeConv for 3D object detection and occupancy.

This paper represents the first application of the \emph{Shape} notion to 2D tasks. 
In particular, we observe that the moiré patterns tend to be repetitive across channel, aligning with the features that \emph{Shape} learns.
Additionally, we propose an application of \emph{Shape} at the global image-level, namely the Shape-Architecture, to complement the patch-level ShapeConv.
Our experimental results demonstrate a distinct advantage of \emph{Shape} when the input data changes from 3D to 2D.
This showcases the versatility of the \emph{Shape} concept and holds promise for broader applications.

\section{Method}
\label{sec:method}

\begin{figure*}[t!]
	\centering
	\centering
	\includegraphics[width=0.98\textwidth]{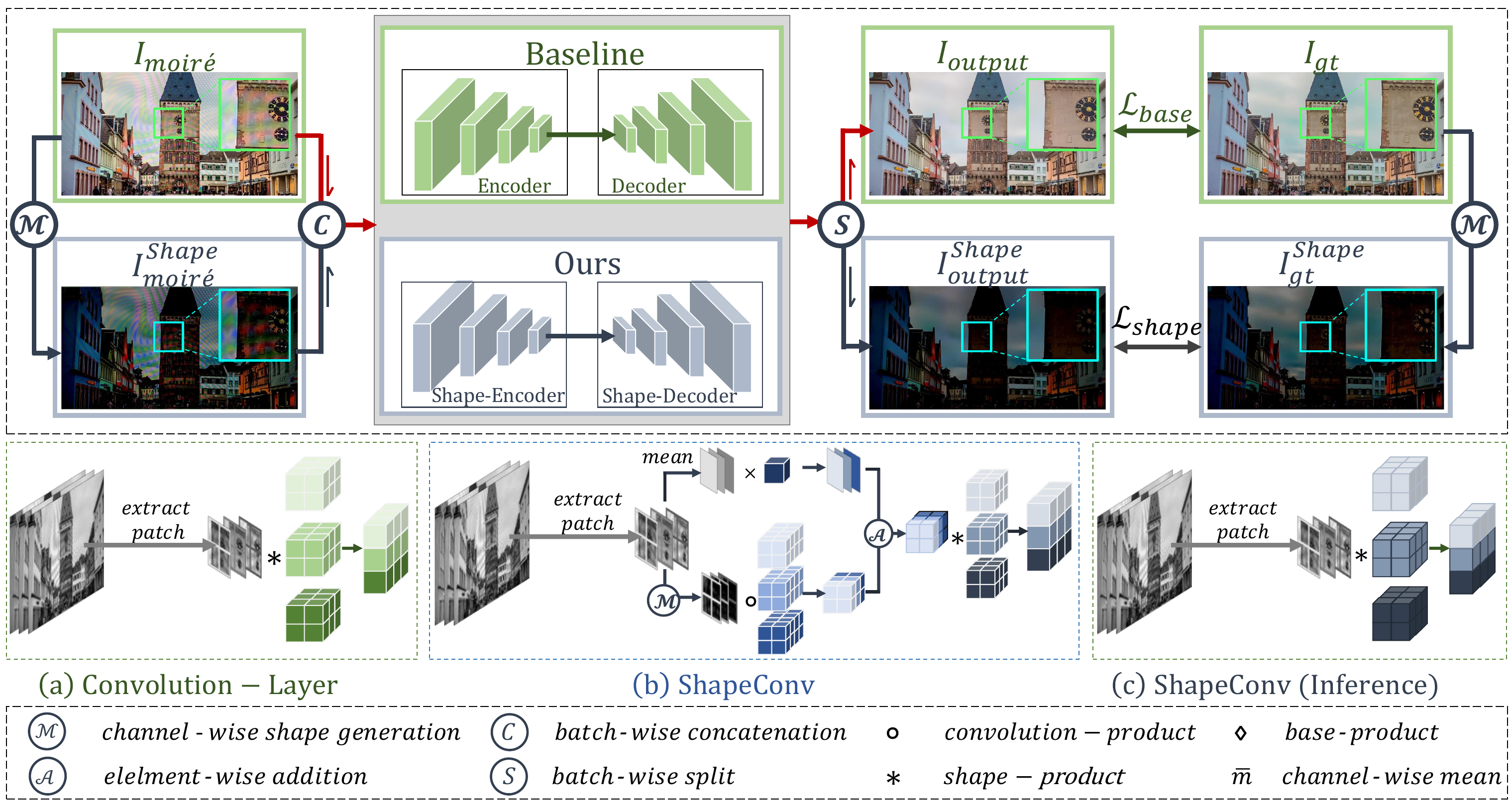}
 
	\caption{
 Overview of ShapeMoiré. The top part depicts the network's overall architecture. The \textcolor{green}{green} and \textcolor{blue}{blue} parts denote the baseline and ShapeMoiré components, respectively. and The \textcolor{red}{red} line flow represents ShapeMoiré during inference. The bottom part shows network layers: (a) Convolution-Layer (consistent in both training and testing), and (b) and (c) illustrate ShapeConv during training and inference, respectively. For both ShapeConv and Shape-Architecture, our method introduces no parameters and negligible computational cost during inference.
 }
	\label{fig:pipeline}
 \end{figure*}

In this section, we first provide the basic formulation of the Shape-aware convolutional layer (ShapeConv), followed by its application in the training and inference phase. Subsequently, we delve into the comprehensive framework of the Shape-aware network architecture (Shape-Architecture), elucidating its data flow dynamics during both training and inference stages. Finally, we present the ShapeMoiré method, equipped with ShapeConv and Shape-Architecture.

\subsection{ShapeConv}

\paragraph{Method Intuition}

Given an input image with a moiré pattern, $I_{moir\acute{e}}$, image demoiréing aims to remove these unwanted patterns and thereby restore the original pristine image. 
Existing methods~\cite{liu2020wavelet,yu2022towards,he2020fhde,sun2018moire} often involve an encoder-decoder framework, wherein the conventional convolutional layer serves as the core building block. 
A convolutional layer takes as input a feature map, either raw pixels or intermediate image features, in a sliding window manner.
It typically comprises multiple convolution kernels, which are applied to a patch of images at each window position. 
To facilitate our discussion, we conceptualize one image patch as a 3D tensor $\mathbb{P} \in R^{K_h \times K_w \times C_{in}}$, $K_h$ and $K_w$ are the spatial dimensions of the kernel and $C_{in}$ represents the channel numbers in the input feature map. The output features from the vanilla convolution layer are obtained by,
\begin{equation}
\label{eq:conv}
    \mathbb{F} = Conv(\mathbb{K}, \mathbb{P}),  
\end{equation}
where $\mathbb{K} \in R^{K_h \times K_w \times C_{in} \times C_{out}}$ denotes the learnable weights of kernels in a convolutional layer (The bias terms are not included for simplicity.) and $C_{out}$ represents the number of channels in the output feature map. Each element of $\mathbb{F} \in R^{C_{out}}$ is calculated as,
\begin{equation}
\begin{cases}
   \mathbb{F} = \mathbb{K} \circ \mathbb{P}\\
    \mathbb{F}_{c_{out}} = \sum_{i}^{K_h \times K_w \times C_{in}} (\mathbb{K}_{i,c_{out}} \times \mathbb{P}_i).
\end{cases}
\label{eq:conv-product}
\end{equation}

where the $\circ$ denotes the convolution-product. We show one example in Fig.~\ref{fig:pipeline}(a) (best view in color), where $K_h=K_w=2$, $C_{in}=3$ (featuring distinct input channels represented by various square frame colors), and $C_{out}=3$.

\paragraph{ShapeConv Formulation}
Based on the aforementioned analysis, in this paper, we offer to decompose an input patch into two components: a base-component $\mathbb{P}_B$ and a shape-component $\mathbb{P}_S$. Specifically, we refer to the mean of patch values to be $\mathbb{P}_B$, and its relative values to be as $\mathbb{P}_S$:
\begin{equation}
   \begin{aligned}
    & \mathbb{P}_B = \overline{m}(\mathbb{P}), \\
    & \mathbb{P}_S = \mathcal{M}(\mathbb{P}) = \mathbb{P} - \overline{m}(\mathbb{P}),
   \end{aligned}
\end{equation}
where $\overline{m}(\mathbb{P})$ is the channel-wise mean function on $\mathbb{P}$ (over the $K_h \times K_w$ dimensions), with $\mathbb{P}_B \in R^{1 \times 1 \times C_{in}}$, and $\mathbb{P}_S \in R^{K_h \times K_w \times C_{in}}$.

Note that directly convolved $\mathbb{P}_S$ with $\mathbb{K}$ in Equation~\ref{eq:conv} is sub-optimal, as the values from $\mathbb{P}_B$ contribute significantly to key discrimination across patches. Therefore, our ShapeConv approach utilizes two learnable weights, $\mathbb{W}_B \in R^{1}$ and $\mathbb{W}_S \in R^{K_h \times K_w \times K_h \times K_w \times C_{in}}$ , to separately process these two components. The resulting features are then combined through element-wise addition, creating a new channel-wise shape-aware patch with the same size as the original $\mathbb{P}$. The formulation of ShapeConv is given as,

\begin{equation}
\label{eq:shapeconv-patch}
\begin{aligned}
 \mathbb{F} &= ShapeConv(\mathbb{K}, \mathbb{W}_B, \mathbb{W}_S, \mathbb{P})\\
              &= Conv(\mathbb{K}, \mathbb{W}_B \diamond \mathbb{P}_B + \mathbb{W}_S \ast \mathbb{P}_S)\\
              &= Conv(\mathbb{K}, \textbf{P}_\textbf{B} + \textbf{P}_\textbf{S})\\
              &= Conv((\mathbb{K}, \textbf{P}_\textbf{BS}),
\end{aligned}
\end{equation}
where $\diamond$ and $\ast$ denote the base-product and shape-product operator, respectively, which are defined as,
\begin{equation}
\begin{cases}
   \textbf{P}_\textbf{B} = \mathbb{W}_B \diamond \mathbb{P}_B \\
   \textbf{P}_{\textbf{B}_{1,1,c_{in}}} = \mathbb{W}_B \times \mathbb{P}_{B_{1,1,c_{in}}},
\end{cases}
\label{eq:base-product-P}
\end{equation}
\begin{equation}
\begin{cases}
    \textbf{P}_\textbf{S} = \mathbb{W}_S \ast \mathbb{P}_S \\
   \textbf{P}_{\textbf{S}_{{k_h},{k_w},{c_{in}}}} = \sum_{i}^{K_h \times K_w} (\mathbb{W}_{S_{i,{k_h},{k_w},{c_{in}}}} \times \mathbb{P}_{S_{i,{c_{in}}}}),
\end{cases}
\label{eq:shape-product-P}
\end{equation}
where $c_{in}$, $k_h$, $k_w$ are the indices of the elements in $C_{in}$, $K_h$, $K_w$ dimensions, respectively.

We reconstruct the channel-wise shape-aware patch $\textbf{P}_\textbf{BS}$ through the addition of $\textbf{P}_\textbf{B}$ and $\textbf{P}_\textbf{S}$, where $\textbf{P}_\textbf{BS} \in R^{K_h \times K_w \times C_{in}}$, enabling it to be smoothly convolved by the kernel $\mathbb{K}$ of a vanilla convolutional layer.
However, $\textbf{P}_\textbf{BS}$ is equipped with the important shape information which is learned by two additional weights. The operation of the ShapeConv is depicted at the bottom of Fig.~\ref{fig:pipeline}(b).

\subsection{ShapeConv in Training and Inference}
\paragraph{Training phase} The proposed ShapeConv can effective leverage the \emph{shape} information of patches. However, replacing the vanilla convolutional layer with ShapeConv in CNNs introduces more computational cost due to the two \emph{product} operations in Equation~\ref{eq:base-product-P} and ~\ref{eq:shape-product-P}. To tackle this problem, we propose shifting these two operations from patches to kernels,
\begin{equation*}
\begin{aligned}
\begin{cases}
    \textbf{K}_\textbf{B} = \mathbb{W}_B \diamond \mathbb{K}_B \\
   \textbf{K}_{\textbf{B}_{1,1,c_{in},c_{out}}} = \mathbb{W}_B \times \mathbb{K}_{B_{1,1,c_{in},c_{out}}},
\end{cases}
\label{eq:base-product-K}
\end{aligned}
\end{equation*}
\begin{equation*}
\begin{aligned}
\begin{cases}
     \textbf{K}_\textbf{S} = \mathbb{W}_S \ast \mathbb{K}_S \\
   \textbf{K}_{\textbf{S}_{{k_h},{k_w},{c_{in}},{c_{out}}}} = \sum_{i}^{K_h \times K_w} (\mathbb{W}_{S_{i,{k_h},{k_w},{c_{in}}}} \times \mathbb{K}_{S_{i,{c_{in}},{c_{out}}}}),
\end{cases}
\label{eq:shape-product-K}
\end{aligned}
\end{equation*}
where $\mathbb{K}_B \in R^{1 \times 1 \times C_{in} \times C_{out}}$ and $\mathbb{K}_S \in R^{K_h \times K_w \times C_{in} \times C_{out}}$ denote the base-component of kernels and shape-component, respectively, and $\mathbb{K} = \mathbb{K}_B + \mathbb{K}_S$. 

We therefore re-formalize ShapeConv in Equation~\ref{eq:shapeconv-patch} to:
\begin{equation}
\label{eq:shapeconv-kernel}
\begin{aligned}
 \mathbb{F} &= ShapeConv(\mathbb{K}, \mathbb{W}_B, \mathbb{W}_S, \mathbb{P})\\
 &= Conv(\mathbb{W}_B \diamond \overline{m}(\mathbb{K})+ \mathbb{W}_S \ast \mathcal{M}(\mathbb{K}), \mathbb{P})\\
              &= Conv(\mathbb{W}_B \diamond \overline{m}(\mathbb{K})+ \mathbb{W}_S \ast (\mathbb{K} - \overline{m}(\mathbb{K})), \mathbb{P})\\
              &= Conv(\mathbb{W}_B \diamond \mathbb{K}_B+ \mathbb{W}_S \ast \mathbb{K}_S, \mathbb{P})\\
              &= Conv(\textbf{K}_\textbf{B} + \textbf{K}_\textbf{S}, \mathbb{P})\\
              &= Conv(\textbf{K}_\textbf{BS}, \mathbb{P}),
\end{aligned}
\end{equation}
where $\overline{m}(\mathbb{K})$ is the channel-wise mean function on $\mathbb{K}$ (over the $K_h \times K_w$ dimensions). And we require $\textbf{K}_\textbf{BS} = \textbf{K}_\textbf{B} + \textbf{K}_\textbf{S}$, $\textbf{K}_\textbf{BS} \in R^{K_h \times K_w \times C_{in} \times C_{out}}$.

\begin{figure}[t!]
	\centering
	\includegraphics[width=0.95\textwidth]{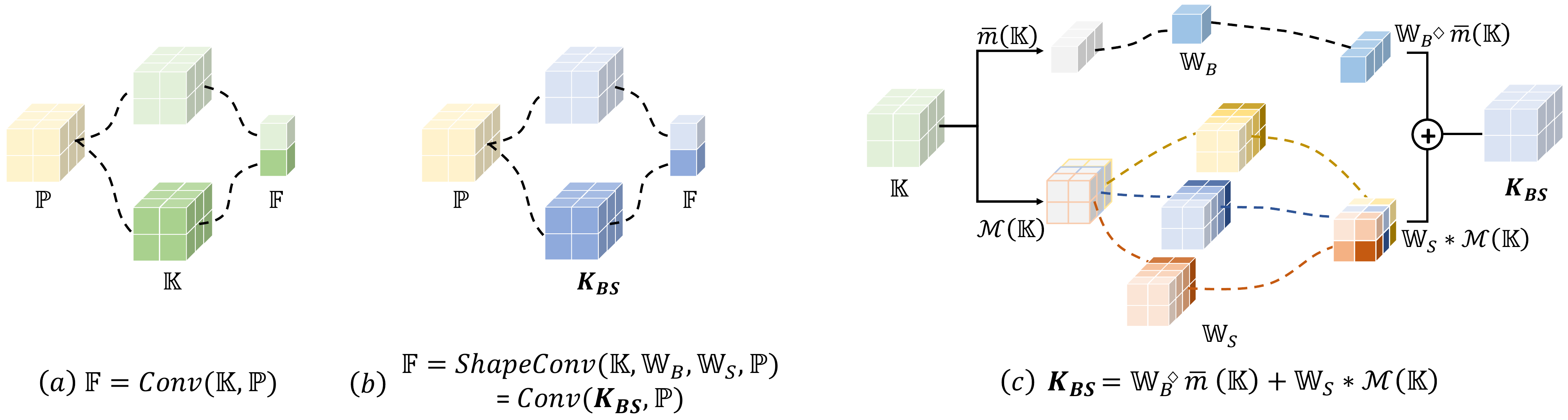}

	\caption{Comparison of vanilla convolution layer and ShapeConv within a patch $\mathbb{P}$. In this figure, $K_h = K_w = 2$, $C_{in} = 3$, and $C_{out} = 2$, ``+'' denotes element-wise addition. (a) Vanilla convolution with kernel $\mathbb{K}$; (b) ShapeConv with folding the $\mathbb{W}_B$ and $\mathbb{W}_S$ into $\textbf{K}_\textbf{BS}$; (c) The computation of $\textbf{K}_\textbf{BS}$ from $\mathbb{K}$, $\mathbb{W}_B$ and $\mathbb{W}_S$.
	}
	\label{fig:ShapeConv}
\end{figure}

In fact, the two formulations of Shpe-Layer, i.e., Equation~\ref{eq:shapeconv-patch} and Equation~\ref{eq:shapeconv-kernel} are mathematically equivalent, i.e.,

\begin{equation}
\begin{aligned}
\mathbb{F} &= ShapeConv(\mathbb{K}, \mathbb{W}_B, \mathbb{W}_S, \mathbb{P})\\
              &= Conv(\mathbb{K}, \textbf{P}_\textbf{BS})\\
              &= Conv(\textbf{K}_\textbf{BS}, \mathbb{P}),
\end{aligned}
\end{equation}

\begin{equation}
\begin{aligned}
     \mathbb{F}_{c_{out}} &= \sum_{i}^{K_h \times K_w \times C_{in}} (\mathbb{K}_{i,{c_{out}}} \times \textbf{P}_{\textbf{BS}_{i}})\\
     &= \sum_{i}^{K_h \times K_w \times C_{in}} (\textbf{K}_{\textbf{BS}_{i,{c_{out}}}} \times \mathbb{P}_{i}),
\end{aligned}
\end{equation}
please refer to the Appendix for detailed proof. In this way, we utilize the ShapeConv in Equation~\ref{eq:shapeconv-kernel} in our implementation as illustrated in Figure~\ref{fig:ShapeConv}(b) and (c).

\paragraph{Inference phase} 
During inference, since the two additional weights i.e. $\mathbb{W}_B$ and $\mathbb{W}_S$, become constants, we can fuse them into $\textbf{K}_\textbf{BS}$ as shown in Figure~\ref{fig:ShapeConv}(c) with $\textbf{K}_\textbf{BS} = \mathbb{W}_B \diamond \mathbb{K}_B+ \mathbb{W}_S \ast \mathbb{K}_S$. 
And $\textbf{K}_\textbf{BS}$ shares the same tensor size with $\mathbb{K}$ in Equation~\ref{eq:conv}, thus, our ShapeConv is actually the same as the vanilla convolutional layer in Figure~\ref{fig:ShapeConv} (a). 
In this way, this operation introduces negligible computational and memory costs compared to the original convolution layer (see Fig.~\ref{fig:pipeline} (a) and (c) for a visual comparison).

\subsection{Shape-Architecture}

\noindent
\paragraph{Method Intuition}
As depicted in Fig.~\ref{fig:pipeline} (top), the baseline network architecture (shown in green boxes) takes the image with moiré patterns, denoted as $I_{moir\acute{e}}$, as input and produces the restoration image, denoted as$I_{output}$. Subsequently, the $\mathcal{L}_{base}$ is computed by comparing $I_{output}$ with the ground truth, represented as $I_{gt}$. The network parameters can then be updated using typical gradient descent algorithms.
Current prevailing methods commonly utilize pixel-wise loss functions, such as $L_1$ loss~\cite{wang2021image} and $L_2$ loss~\cite{sun2018moire}, or a combination with the feature-based perceptual loss $L_p$~\cite{liu2020wavelet,yu2022towards,he2019mop,guo2021loss}. In this context, we take the combination loss as an example,
 \begin{equation}
\mathcal{L}_{base} = L_1(I_{output}, I_{gt})+\lambda L_p(I_{output}, I_{gt}),
\end{equation}
where $\lambda$ is a hyperparameter that balances between $L_1$ and $L_p$.

\paragraph{Shape-Architecture}
 
The above ShapeConv primarily operates on intermediate features.
To further explore the utility of \emph{Shape} features, we leverage a Shape-Architecture to extract these features from the image-level.
The Shape-Architecture complements the baseline network architecture with an additional \emph{Shape} stream, as illustrated in the blue box and solid data flow in Fig.~\ref{fig:pipeline}.
To implement this, we first acquire the shape feature $I^{Shape}_{moir\acute{e}}$ from the input image.
Thereafter, we input $I^{Shape}_{moir\acute{e}}$ to the same model and employ the ground-truth shape features, $I^{Shape}_{gt}$ which are extracted from $I_{gt}$, for supervision, 
\begin{equation}
    \begin{aligned}
I^{Shape}_{moir\acute{e}}&=\mathcal{M}(I_{moir\acute{e}}),\\
I^{Shape}_{gt}&=\mathcal{M}(I_{gt}),
\end{aligned}
\end{equation}
where $\mathcal{M}$ aligns with the definition in ShapeConv, i.e., the subtraction from the channel-wise mean. Subsequently, $I^{Shape}_{moir\acute{e}}$ is batch-wisely concatenated with $I_{moir\acute{e}}$ and fed into the model, ensuring that these two streams remain uninterfered with each other.
The model output is then split into $I_{output}$ and $I^{Shape}_{output}$.
Pertaining to the shape stream, the loss function $\mathcal{L}_{shape}$ is obtained by,
 \begin{equation}
 \begin{aligned}
\mathcal{L}_{shape} = L_1(I^{Shape}_{output}, I^{Shape}_{gt}) +\lambda L_p(I^{Shape}_{output}, I^{Shape}_{gt}).
\end{aligned}
\end{equation}
The final loss function $\mathcal{L}_{total}$ is:
\jm{\begin{equation}
\mathcal{L}_{total} = \alpha\mathcal{L}_{base} + \beta\mathcal{L}_{shape},
\end{equation}}
\jm{where $\alpha$ and $\beta$ are the weighting coefficients for the loss terms $\mathcal{L}_{base}$ and $\mathcal{L}_{shape}$, respectively.}

\paragraph{Shape-Architecture During Inference}
Note that during training, the two streams share the same parameters.
As a result, no parameters are introduced by our Shape-Architecture.
We leverage $\mathcal{L}_{shape}$ to make the parameters more aware of the shape features.
Regarding the inference stage, in order to align with the structure of the baseline network architecture, we efficiently remove the shape stream from our Shape-Architecture, as illustrated in the red line flow in Fig.~\ref{fig:pipeline}.
Moreover, we employ this strategy to allow for seamless integration with baselines and avoid introducing any computational overheads.

\jm{\paragraph{Relationship between ShapeConv and Shape-Architecture.}
While ShapeConv and Shape-Architecture operate at different levels—patch-level and image-level respectively—they are intrinsically aligned through the shared notion of shape decomposition. Specifically, both components rely on the same shape operator $\mathcal{M}$ to extract the relative information from input data by subtracting the channel-wise mean. ShapeConv uses this operator within local patches to enhance feature discrimination, whereas Shape-Architecture applies it globally to form a parallel shape stream at the image level. This design ensures consistency in how shape information is perceived and utilized throughout the network. Together, they form a hierarchical shape-aware framework, where ShapeConv captures local fine-grained shape variations and Shape-Architecture reinforces the model’s global structural awareness. This joint modeling helps the network better suppress moiré patterns that exhibit both local distortions and large-scale repetitive interference.}

\subsection{ShapeMoiré}
We term the approach equipped with both ShapeConv and Shape-Architecture as ShapeMoiré. 
This means our ShapeMore operates on two hierarchical levels:
1) Patch-level: we replace all the convolutional layers with our ShapeConv;
2) Image-level: we complement the baseline network architecture with another shape-aware image stream.
It is noteworthy that the two components, ShapeConv and Shape-Architecture, are independent of each other and can be separately added to baseline methods to enhance performance. As mentioned earlier, neither of these components introduces additional parameters or computational overhead during inference. Therefore, our approaches introduce ZERO parameters and incur negligible computation compared to the baseline method during inference.
We will later demonstrate the complementary effect of these two components in our experiments.
\begin{table*}[]
\centering
\caption{\label{tab:sota}
	Performance comparison with state-of-the-art methods on LCDMoiré, TIP2018, FHDMi, and UHDM datasets. ($\uparrow$) and ($\downarrow$) indicate better performance with larger and smaller numbers, respectively. \textcolor{red}{Red}: best and \textcolor{blue}{Blue}: second-best.
 }
\scalebox{0.7}{
\centering
\begin{tabular}{c|c|ccccccccc
>{\columncolor[HTML]{D0CECE}}c }
\hline
                                                                                                               &                           &                         & DMCNN                                & MopNet                            & WDNet                                  & MBCNN                                  & ESDNet                                & SEDNet-L                              & DDA                                   & RRID                                & \cellcolor[HTML]{D0CECE}                             \\
                                                                                                               &                           &                         & (TIP18)                              & (ICCV19)                          & (ECCV20)                               & (CVPR20)                               & (ECCV22)                              & (ECCV22)                              & (ICLR23)                              & (arXiv24)                           & \cellcolor[HTML]{D0CECE}                             \\
\multirow{-3}{*}{Dataset}                                                                                      & \multirow{-3}{*}{Metrics} & \multirow{-3}{*}{Input} & ~\cite{sun2018moire} &~\cite{he2019mop} & ~\cite{liu2020wavelet} & ~\cite{zheng2020image} & ~\cite{yu2022towards} & ~\cite{yu2022towards} & ~\cite{zhang2023real} & ~\cite{xu2023image} & \multirow{-3}{*}{\cellcolor[HTML]{D0CECE}ShapeMoiré} \\ \hline
                                                                                                         \multirow{-1}{*}{LCDMoiré}      & PSNR$\uparrow$            & 10.44                   & 35.48                                & -                                 & 29.66                                  & 44.04                                  & 44.83                                 & \color[HTML]{3166FF}45.34                                 & 41.68                                 & -                                   & \textcolor{red}{46.56}                                                \\
~\cite{yuan2019aim} & SSIM$\uparrow$            & .5717                  & .9785                               & -                                 & .9670                                  & .9948                                 & \color[HTML]{3166FF}.9963                                &\color[HTML]{FE0000} .9966                               & .9869                                & -                                   & .9930                                                \\ \hline
                                                                                                              TIP2018 & PSNR$\uparrow$            & 20.3                    & 26.77                                & 27.75                             & 28.08                                  & 30.03                                  & 29.81                                 & \color[HTML]{3166FF}30.11                                 & -                                     & -                                   &\color[HTML]{FE0000} 30.16                                               \\
~\cite{sun2018moire} & SSIM$\uparrow$            & .7380                   & .8710                                & .8950                             & .9040                                  & .8930                                  & \color[HTML]{3166FF}.9160                                 &\color[HTML]{FE0000} .9200                                  & -                                     & -                                   & .8795                                               \\ \hline
                                                                                                               & PSNR$\uparrow$            & 17.97                   & 21.54                                & 22.76                             & -                                      & 22.31                                  & 24.5                                  & \color[HTML]{3166FF}24.88                                 & 23.62                                 & 24.39                               &\color[HTML]{FE0000} 25.06                                                \\
                                                                                                      \multirow{-2}{*}{FHDMi  }             & SSIM$\uparrow$            & .7033                  & .7727                               & .7958                            & -                                      & .8095                                 & .8351                                & \color[HTML]{3166FF}.8440                                 & .8293                                & .8300                                &\color[HTML]{FE0000} .8474                                               \\
    ~\cite{he2020fhde}    & LPIPS$\downarrow$         & .2837                  & .2477                               & .1794                            & -                                      & .1980                                  & .1354                                &\color[HTML]{FE0000} .1301                                & -                                     & -                                   & \color[HTML]{3166FF}.1306                                               \\ \hline
                                                                                                               & PSNR$\uparrow$            & 17.12                   & 19.91                                & 19.49                             & 20.36                                  & 21.41                                  & 22.12                                 & \color[HTML]{3166FF}22.42                                 & -                                     & -                                   &\color[HTML]{FE0000} 22.95                                                \\
                                                                                                  \multirow{-2}{*}{UHDM~}             & SSIM$\uparrow$            & .5089                  & .7575                               & .7572                            & .6497                                 & .7932                                 & .7956                                & \color[HTML]{3166FF}.7985                                & -                                     & -                                   &\color[HTML]{FE0000} .8021                                              \\
~\cite{yu2022towards}   & LPIPS$\downarrow$         & .5314                  & .3764                               & .3857                            & .4882                                 & .3318                                 & .2551                                & \color[HTML]{3166FF}.2454                                & -                                     & -                                   &\textcolor{red}{ .2415   }                                            \\ \hline
\end{tabular}
}
\end{table*}

\section{Experiments}
\label{sec:ex}

\noindent
\textbf{Datasets.}
Our experiments were conducted using four publicly accessible datasets:

(1) LCDMoiré Dataset~\cite{yuan2019aim}. As a part of the AIM19 image demoiréing challenge, the LCDMoiré dataset comprises 10,200 synthetically generated image pairs. It is structured to include 10,000 images for training purposes, 100 images for validation, and another 100 for testing. Each image is of high resolution, with dimensions of 1024 $\times$ 1024 pixels. 

(2) TIP2018 Dataset~\cite{sun2018moire}. This dataset consists of real photographs obtained by capturing images from ImageNet~\cite{deng2009imagenet} displayed on computer screens, ensuring each image has a resolution of 256 $\times$ 256 pixels. 

(3) FHDMi Dataset~\cite{he2020fhde}. It includes a substantial collection of 9,981 training image pairs and 2,019 testing pairs. All images in this dataset are in full high-definition resolution, measuring 1920 $\times$ 1080 pixels.

(4) UHDM Dataset~\cite{yu2022towards}. It stands out for its 4K resolution demoiréing capability. It is an ultra-high-resolution dataset featuring 5,000 image pairs that span a broad spectrum of scenes, including landscapes, sports, video clips, and documents. The dataset is notable for its diverse moiré patterns, produced using various device combinations and viewpoints. Each image in this dataset is 3840 $\times$ 2160 pixels.

These datasets, ranging from synthetically generated pairs to ultra-high-resolution real photographs, offer a comprehensive platform for evaluating the effectiveness of our methods across different scenarios and image qualities.

\noindent
\textbf{Evaluation Protocols.}
For quantitative evaluation, we employed widely recognized metrics, including PSNR (Peak Signal-to-Noise Ratio), SSIM (Structure Similarity)~\cite{wang2004image}, and LPIPS (Learned Perceptual Image Patch Similarity)~\cite{zhang2018unreasonable}. 
In addition to these performance-related metrics, we also considered the network's parameter count, denoted as Param.(M), which indicates the number of network parameters in millions. This metric is pertinent to the overall memory usage of the network.
It is important to note that previous methods~\cite{sun2018moire,cheng2019multi,he2019mop,liu2020wavelet,yu2022towards} have exclusively reported PSNR and SSIM metrics on the TIP2018 and LCDMoiré datasets. In line with this convention, we adhere to the same setup for our comparisons.
\begin{table*}[]
\caption{\label{tab:baseline}
	Performance comparison with different baseline methods on four datasets. An asterisk (*) indicates the performance we reproduced on the respective datasets using the source code provided by different methods, noting slight discrepancies from the original publications.
 }
 \centering
\scalebox{0.7}{
\begin{tabular}{c|c|ccc|ccc|cc|cc|c}
\hline
                                                                                                    &                           & \multicolumn{3}{c|}{UHDM~\cite{yu2022towards}}                                                                        & \multicolumn{3}{c|}{FHDMi~\cite{he2020fhde}}                                                                       & \multicolumn{2}{c|}{TIP2018~\cite{sun2018moire}}                                   & \multicolumn{2}{c|}{LCDMoiré~\cite{yuan2019aim}}                                  &                                                                         \\ \cline{3-12}
\multirow{-2}{*}{\begin{tabular}[c]{@{}c@{}}Archi-\\ tecture\end{tabular}}                          & \multirow{-2}{*}{Method}  & PSNR$\uparrow$                     & SSIM$\uparrow$                         & LPIPS $\downarrow$                          & PSNR$\uparrow$                          & SSIM  $\uparrow$                         & LPIPS$\downarrow$                           & PSNR $\uparrow$                         & SSIM $\uparrow$                          & PSNR $\uparrow$                         & SSIM$\uparrow$                           & \multirow{-2}{*}{\begin{tabular}[c]{@{}c@{}}Params.\\ (M)\end{tabular}} \\ \hline
                                                                                                    & Baseline*                  & 22.253                        & .7974                         & .2511                          & 24.393                        & .8392                         & .1366                          & 29.791                        & .8753                         & 45.286                        & .9921                         & 5.934                                                                   \\
                                                                                                    & ShapeMoiré                & 22.597                        & .8007                         & .2490                           & 24.629                        & .8402                         & .1336                          & 29.862                        & .8758                         & 45.537                        & .9925                         & 5.934                                                                   \\
\multirow{-3}{*}{\begin{tabular}[c]{@{}c@{}}ESDNet\\ \cite{yu2022towards}\end{tabular}}   & \cellcolor[HTML]{D0CECE}$\Delta$ & \cellcolor[HTML]{D0CECE}0.344$\uparrow$ & \cellcolor[HTML]{D0CECE}.0033$\uparrow$ & \cellcolor[HTML]{D0CECE}.0021$\downarrow$ & \cellcolor[HTML]{D0CECE}0.236$\uparrow$ & \cellcolor[HTML]{D0CECE}.0010$\uparrow$  & \cellcolor[HTML]{D0CECE}.0030$\downarrow$  & \cellcolor[HTML]{D0CECE}0.071$\uparrow$ & \cellcolor[HTML]{D0CECE}.0005$\uparrow$ & \cellcolor[HTML]{D0CECE}0.251$\uparrow$ & \cellcolor[HTML]{D0CECE}.0004$\uparrow$ & \cellcolor[HTML]{D0CECE}0                                               \\ \hline
                                                                                                    & Baseline*                  & 22.554                        & .7997                         & .2468                          & 24.808                        & .8435                         & .1321                          & 30.096                        & .8789                         & 45.544                        & .9925                         & 10.623                                                                  \\
                                                                                                    & ShapeMoiré                & 22.948                        & .8021                         & .2415                          & 25.064                        & .8474                         & .1306                          & 30.161                        & .8795                         & 46.558                        & .993                          & 10.623                                                                  \\
\multirow{-3}{*}{\begin{tabular}[c]{@{}c@{}}ESDNet-L\\ \cite{yu2022towards}\end{tabular}} & \cellcolor[HTML]{D0CECE}$\Delta$ & \cellcolor[HTML]{D0CECE}0.394$\uparrow$ & \cellcolor[HTML]{D0CECE}.0024$\uparrow$ & \cellcolor[HTML]{D0CECE}.0053$\downarrow$ & \cellcolor[HTML]{D0CECE}0.256$\uparrow$ & \cellcolor[HTML]{D0CECE}.0039$\uparrow$ & \cellcolor[HTML]{D0CECE}.0015$\downarrow$ & \cellcolor[HTML]{D0CECE}0.065$\uparrow$ & \cellcolor[HTML]{D0CECE}.0006$\uparrow$ & \cellcolor[HTML]{D0CECE}1.014$\uparrow$ & \cellcolor[HTML]{D0CECE}.0005$\uparrow$ & \cellcolor[HTML]{D0CECE}0                                               \\ \hline
                                                                                                    & Baseline*                  & 19.181                        & .6417                         & .4271                          & 21.161                        & .7714                         & .2411                          & 27.812                        & .8381                         & 37.324                        & .9721                         & 3.36                                                                    \\
                                                                                                    & ShapeMoiré                & 19.882                        & .7442                         & .3272                          & 22.182                        & .7901                         & .2134                          & 28.312                        & .8535                         & 38.408                        & .9782                         & 3.36                                                                    \\
\multirow{-3}{*}{\begin{tabular}[c]{@{}c@{}}WDNet\\ \cite{liu2020wavelet}\end{tabular}}    & \cellcolor[HTML]{D0CECE}$\Delta$ & \cellcolor[HTML]{D0CECE}0.701$\uparrow$ & \cellcolor[HTML]{D0CECE}.1025$\uparrow$ & \cellcolor[HTML]{D0CECE}.0999$\downarrow$ & \cellcolor[HTML]{D0CECE}1.021$\uparrow$ & \cellcolor[HTML]{D0CECE}.0187$\uparrow$ & \cellcolor[HTML]{D0CECE}.0277$\downarrow$ & \cellcolor[HTML]{D0CECE}0.500$\uparrow$   & \cellcolor[HTML]{D0CECE}.0154$\uparrow$ & \cellcolor[HTML]{D0CECE}1.084$\uparrow$ & \cellcolor[HTML]{D0CECE}.0061$\uparrow$ & \cellcolor[HTML]{D0CECE}0                                               \\ \hline
                                                                                                    & Baseline*                  & 17.812                        & .7294                         & .5561                          & 19.313                        & .7375                         & .3395                          & 24.518                        & .8073                         & 31.425                        & .9552                         & 1.426                                                                   \\
                                                                                                    & ShapeMoiré                & 18.036                        & .7336                         & .5185                          & 19.615                        & .7495                         & .2889                          & 25.381                        & .8234                         & 31.789                        & .9562                          & 1.426                                                                   \\
\multirow{-3}{*}{\begin{tabular}[c]{@{}c@{}}DMCNN\\ \cite{sun2018moire} \end{tabular}}    & \cellcolor[HTML]{D0CECE}$\Delta$ & \cellcolor[HTML]{D0CECE}0.224$\uparrow$ & \cellcolor[HTML]{D0CECE}.0042$\uparrow$ & \cellcolor[HTML]{D0CECE}.0376$\downarrow$ & \cellcolor[HTML]{D0CECE}0.302$\uparrow$ & \cellcolor[HTML]{D0CECE}.0120$\uparrow$  & \cellcolor[HTML]{D0CECE}.0506$\downarrow$ & \cellcolor[HTML]{D0CECE}0.863$\uparrow$ & \cellcolor[HTML]{D0CECE}.0161$\uparrow$ & \cellcolor[HTML]{D0CECE}0.364$\uparrow$ & \cellcolor[HTML]{D0CECE}.0010$\uparrow$ & \cellcolor[HTML]{D0CECE}0                                               \\ \hline
\end{tabular}
}
\end{table*}

\noindent
\textbf{Implementation Details.}
We adopted several popular approaches~\cite{sun2018moire,liu2020wavelet,yu2022towards} as our baseline to demonstrate both the effectiveness and generalization capability of our proposed method.
We strictly followed the original implementations of these baseline methods, except for the model components of ShapeMoiré. 
This guarantees that the observed performance improvements are solely attributable to the application of ShapeMoiré, yet not influenced by other factors. Unless otherwise noted, the baseline model is ESDNet-L~\cite{yu2022towards}. Detailed implementations of our method can be found in our released code.

\subsection{Comparison with State-of-the-Art Methods}
We compared our method with state-of-the-art techniques and reported the results in Table~\ref{tab:sota}. As evident from the table, our ShapeMoiré significantly outperforms existing methods across most scenarios.
It is noteworthy that our approach establishes a new SOTA across all four datasets in terms of the PSNR metric, which quantifies the level of noise or distortion compared to the original undistorted image. 
The results highlight that the modeled shape features in our Shapemoiré play a crucial role in identifying the causal factors essential for image demoiréing.
Furthermore, ShapeMoiré demonstrates superior performance compared to existing approaches on the UHDM dataset, emphasizing the effectiveness of our proposed method for Ultra-High-Definition images.

\subsection{Generalization to Diverse Architectures}
The proposed ShapeMoiré is a versatile approach that can be seamlessly integrated into existing methods for image demoiréing. 
To test its generalization capability, we applied our ShapeMoiré to four representative demoiréing methods, namely ECDNet~\cite{yu2022towards}, ECDNet-L~\cite{yu2022towards}, WDnet~\cite{liu2020wavelet}, and DMCNN~\cite{sun2018moire}. Table~\ref{tab:baseline} demonstrates that ShapeMoiré consistently delivers significant performance improvements across ALL settings. 
Moreover, our method introduces no additional inference cost compared to baselines, i.e., the added parameters are zero.
It is worth noting that, as observed in Table~\ref{tab:baseline}, the performance of current methods mostly correlates positively with the number of parameters. However, ShapeMoiré demonstrates the ability to enhance performance without increasing the network's parameter count. For instance, our method achieves comparable (or even higher PSNR) performance to ESDNet-L (which has almost twice the parameter count of ESDNet) on the UHDM dataset, even when keeping the parameter count consistent with ESDNet.

\subsection{\jm{Generalization to Image Deblurring}}
\begin{table}[h!]
\caption{\label{tab:deblur}
\jm{Dual-pixel Defocus Deblurring comparisons on the DPDD Dataset~\cite{abuolaim2020defocus}. The test set of DPDD contains 37 indoor
scenes and 39 outdoor scenes.}
 }
 \centering
\scalebox{0.8}{
\begin{tabular}{c|c|ccc|ccc|ccc}
\hline
Archi-                     & \multirow{2}{*}{Method}          & \multicolumn{3}{c|}{Indoor Scenes}                                  & \multicolumn{3}{c|}{Outdoor Scenes}                                 & \multicolumn{3}{c}{Combined}                                       \\ \cline{3-11} 
tecture                    &                                  & PSNR$\uparrow$       & SSIM$\uparrow$       & LPIPS $\downarrow$    & PSNR$\uparrow$       & SSIM$\uparrow$       & LPIPS $\downarrow$    & PSNR$\uparrow$       & SSIM$\uparrow$       & LPIPS $\downarrow$   \\ \hline
\multirow{3}{*}{MIRNet-v2~\cite{zamir2022learning}} & Baseline                         &       28.95               &           .8800           &        .1585               &          23.66            &           .7549           &           .2109            &           26.23           &         .8158             &            .1854          \\
                           & \multicolumn{1}{c|}{ShapreMoiré} & \multicolumn{1}{c}{29.05} & \multicolumn{1}{c}{.8841} & \multicolumn{1}{c|}{.1507} & \multicolumn{1}{c}{23.77} & \multicolumn{1}{c}{.7569} & \multicolumn{1}{c|}{.1974} & \multicolumn{1}{c}{26.34} & \multicolumn{1}{c}{.8188} & \multicolumn{1}{c}{.1747} \\ 
                           & \cellcolor[HTML]{D0CECE}$\Delta$ & \cellcolor[HTML]{D0CECE}0.10$\uparrow$ & \cellcolor[HTML]{D0CECE}.0041$\uparrow$ & \cellcolor[HTML]{D0CECE}.0078$\downarrow$ & \cellcolor[HTML]{D0CECE}0.11$\uparrow$ & \cellcolor[HTML]{D0CECE}.020$\uparrow$  & \cellcolor[HTML]{D0CECE}.0135$\downarrow$ & \cellcolor[HTML]{D0CECE}0.11$\uparrow$ & \cellcolor[HTML]{D0CECE}.0030$\uparrow$ & \cellcolor[HTML]{D0CECE}.0107$\downarrow$               \\ \hline
\end{tabular}}
\end{table}

\jm{To further evaluate the generalization ability of the proposed method beyond the demoiréing task, we extend our experiments to the dual-pixel defocus deblurring task using the DPDD dataset~\cite{abuolaim2020defocus}. We adopt MIRNet-v2~\cite{zamir2022learning} as the baseline architecture and integrate our proposed ShapeConv and Shape-Architecture modules. As presented in Table~\ref{tab:deblur}, the ShapeMoiré-enhanced model consistently outperforms the baseline across all three metrics—PSNR, SSIM, and LPIPS—on both indoor and outdoor subsets. Specifically, on the combined test set, our method improves PSNR from 26.23 to 26.34, SSIM from .8158 to .8188, and reduces LPIPS from .1854 to .1747. These improvements indicate not only better pixel-level fidelity but also enhanced perceptual quality.

The consistent performance gains across different domains and evaluation metrics suggest that ShapeMoiré is a versatile and effective enhancement module that can be generalized to broader image restoration tasks beyond demoiréing.}

\subsection{Ablation Study}

The two components we propose, ShapeConv and Shape-Architecture, are independent of each other. This means that adding either one of them to the baseline method alone can enhance the model's performance. To validate the effectiveness of both components, we conducted detailed ablation studies. The results of these studies are presented in Table~\ref{tab:ablation}. One can observe that the removal of both ShapeConv and Shape-Architecture leads to degraded model performance.
Introducing either the ShapeConv or the Shape-Architecture yields a positive outcome.
Combining these two together, i.e., our final ShapeMoiré, achieves the best results, highlighting their complementary effect on the overall model performance. 
\begin{table}[h!]
\centering
\caption{\label{tab:ablation}
	Ablation study of Shapremoiré on the UHDM dataset. 
 }
\scalebox{0.85}{
\begin{tabular}{cc|ccc}
\hline
ShapeConv & Shape-Architecture & PSNR$\uparrow$            & SSIM$\uparrow$         & LPIPS $\downarrow$     \\ \hline
    $\times$   &   $\times$                 & 22.253          & .7974     &.2511     \\ \hline
$\checkmark$           &   $\times$                 & 22.469         & .7990       &\textbf{.2424}    \\ \hline
      $\times$      & $\checkmark$                  & 22.315         & .7984     &.2471     \\ \hline
$\checkmark$          & $\checkmark$                  & \textbf{22.597} & \textbf{.8007}  &.2490\\ \hline
\end{tabular}
}
\end{table}

Additionally, we computed the inference time of both the baseline and our ShapeMoiré method.
Specifically, we conducted this experiment on a single NVIDIA A40 GPU with 48G memory and ran both models for three runs.
Subsequently, we calculated the average inference time for 5,000 images and presented the milliseconds (ms) per image in Table~\ref{tab:time}.
It is evident that our ShapeMoiré method imposes a very negligible overhead on the baseline model in most cases and even runs faster than the baseline due to clock time oscillation. 
This results validate the efficiency of the proposed ShapeMoiré method.

\begin{table}[h!]
\centering
\caption{\label{tab:time}
 Inference time comparison of baseline and our ShapeMoiré on the UHDM dataset.
 }
\scalebox{0.85}{
\begin{tabular}{c|c|ccc}
\hline
Archi-  &                                    & \multicolumn{3}{c}{ms/image}                                \\ \cline{3-5} 
tecture & \multirow{-2}{*}{Method}           & run1                     & run2                     & run3                     \\ \hline
ESDNet  & Baseline                           & 7.70                     & 8.00                     & 7.80                     \\
\cite{yu2022towards}  & \cellcolor[HTML]{D0CECE}ShapeMoiré & \cellcolor[HTML]{D0CECE}8.30 & \cellcolor[HTML]{D0CECE} 8.04& \cellcolor[HTML]{D0CECE} 7.98\\ \hline
WDNet   & Baseline                           & 6.79                     & 7.07                     & 6.62                     \\
 \cite{liu2020wavelet} & \cellcolor[HTML]{D0CECE}ShapeMoiré & \cellcolor[HTML]{D0CECE} 6.72& \cellcolor[HTML]{D0CECE}7.01 & \cellcolor[HTML]{D0CECE}7.06 \\ \hline
DMCNN   & Baseline                           & 2.17                     & 1.96                     & 1.91                     \\
\cite{sun2018moire}  & \cellcolor[HTML]{D0CECE}ShapeMoiré & \cellcolor[HTML]{D0CECE}1.95 & \cellcolor[HTML]{D0CECE}2.07 & \cellcolor[HTML]{D0CECE}2.21 \\ \hline
\end{tabular}
}
\end{table}

\jm{We conducted experiments with three different settings of the loss weights ($\alpha, \beta$), as shown in Table~\ref{tab:weight}. The results demonstrate that across all configurations, our method consistently outperforms the baseline, indicating that the proposed framework is not highly sensitive to the specific choice of weight parameters and exhibits robust performance. While there are slight performance variations depending on the weighting (e.g., slightly better LPIPS with $\alpha = 10$), we attribute these differences to potential dataset-specific characteristics. To maintain fairness and reproducibility, we adopt the balanced setting ($\alpha = 1, \beta = 1$) for all experiments reported in the main paper.}

\begin{table}[h!]
\caption{\label{tab:weight}
\jm{Ablation study on the weighting coefficients for the loss terms. Experiments are conducted using DMCNN as the baseline method on the LCDMoiré dataset.}
 }
 \centering
\scalebox{0.9}{
\begin{tabular}{cc|ccc}
\hline
$\alpha$ & $\beta$ & PSNR$\uparrow$ & SSIM$\uparrow$ &  LPIPS $\downarrow$ \\ \hline
\multicolumn{2}{c|}{Baseline} & \multicolumn{1}{c}{31.425} & \multicolumn{1}{c}{.9552} & \multicolumn{1}{c}{0.093}  \\ \hline
0.1      & 1.0     &       31.975         &        .9616           &     0.085                  \\
1.0      & 1.0     &        31.789        &        .9562           &     0.088               \\
10       & 1.0     &       32.077         &           .9609        &    0.070               \\ \hline
\end{tabular}}
\end{table}

\subsection{Visualization}

\begin{figure*}[h!]
	\centering
	\centering
	\includegraphics[width=1.0\textwidth]{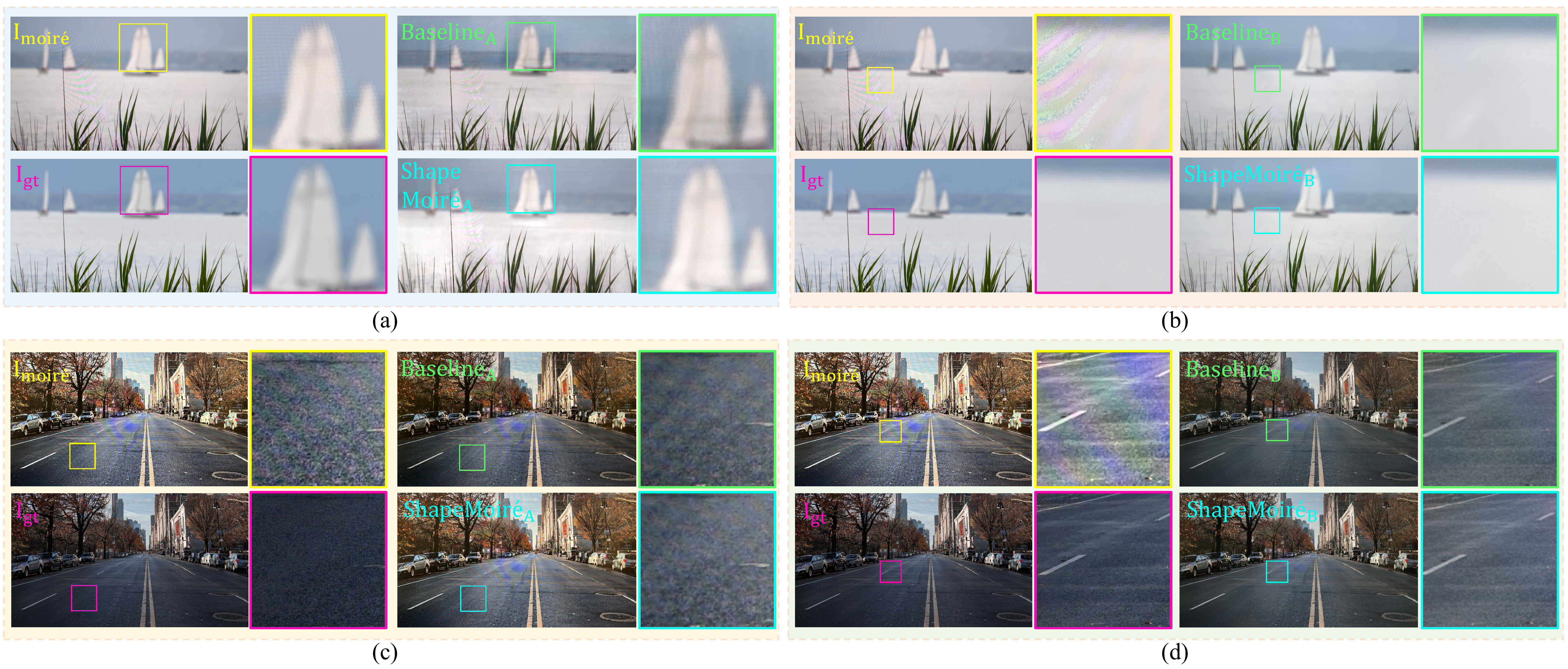}
 
	\caption{\jm{Visualization results from the UHDM dataset.
Subfigures (a), (b), (c), and (d) respectively showcase different sample cases. In each subfigure, the {\color{yellowlight}{$I_{moir\acute{e}}$}}, {\color{pinklight}{$I_{gt}$}}, {\color{green}{$Baseline$}}, and {\color{blue}{\textit{ShapeMoiré}}} correspond to the moiré-contaminated input image, ground truth, baseline output, and ShapeMoiré output, respectively. For each case, $Baseline_{A}$ and \textit{ShapeMoiré}$_{A}$ denote results based on WDNet~\cite{liu2020wavelet}, while $Baseline_{B}$ and \textit{ShapeMoiré}$_{B}$ are derived using ESDNet-L~\cite{yu2022towards} as the baseline. The larger boxes display magnified versions of the corresponding colored boxes, providing a clearer view of fine-grained differences.
 }}
	\label{fig:vis}
 \end{figure*}

\jm{In Fig.~\ref{fig:vis}, we present qualitative comparisons across four sample cases—(a), (b), (c), and (d)—each visualized with moiré-contaminated input, ground truth, baseline output, and our ShapeMoiré output.
We include two representative baselines, WDNet\cite{liu2020wavelet} and ESDNet-L~\cite{yu2022towards}, to assess the generalizability of our method. For each baseline, its corresponding ShapeMoiré variant is denoted as ShapeMoiré$_A$ and ShapeMoiré$_B$, respectively.
First, we observe that stronger baselines (e.g., ESDNet-L) generally produce better restoration results, as seen by comparing the green boxes of (b) and (d) with those of (a) and (c).
However, regardless of the baseline strength, our ShapeMoiré consistently demonstrates superior moiré removal and better structural detail recovery. For instance, in subfigure (a), ShapeMoiré$_A$ (blue box) more accurately reconstructs the sailboat and the surrounding sky compared to the baseline (green box). Similarly, in subfigure (b), ShapeMoiré$_B$ produces more uniform color on the lake surface.
In subfigures (c) and (d), our method further reduces residual moiré artifacts while better preserving textures such as road surfaces.
These observations collectively highlight the robustness and effectiveness of our method across varying baseline strengths and image characteristics.}

\begin{figure*}[h!]
	\centering
	\centering
	\includegraphics[width=1.0\textwidth]{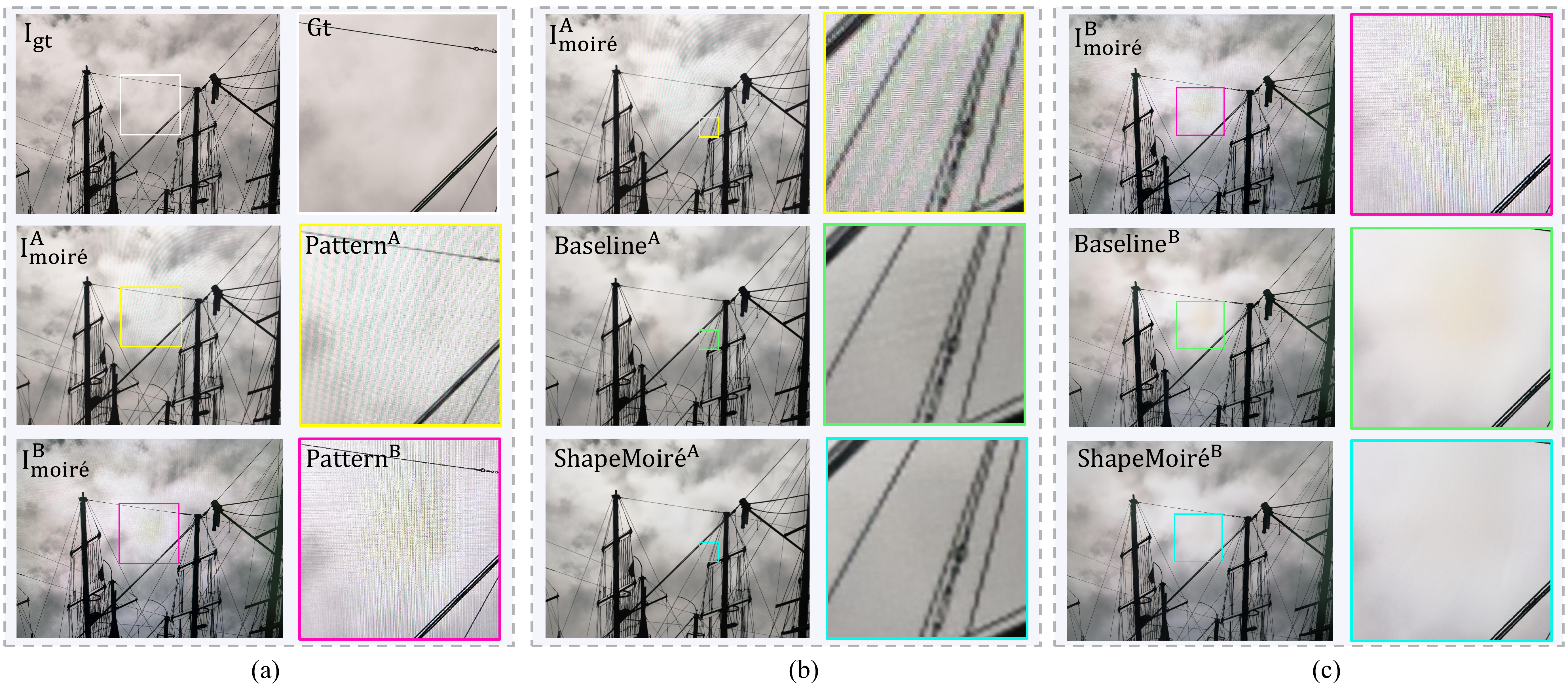}
 
	\caption{\jm{Qualitative results of image demoiréing for real-world cases.
(a) Given the same ground truth image $I_{gt}$, different moiré-contaminated inputs $I^A_{moir\acute{e}}$ and $I^B_{moir\acute{e}}$ are obtained by photographing with different smartphones.
(b) and (c) show the demoiréing results for $I^A_{moir\acute{e}}$ and $I^B_{moir\acute{e}}$ respectively, comparing a baseline method and our proposed ShapeMoiré.
Larger boxes represent enlarged views of the corresponding colored small boxes at the borders, providing a clearer visualization of the moiré removal effectiveness.}
 }
	\label{fig:vis-real}
 \end{figure*}

To further validate the practicality of our proposed method, we captured real-world images using smartphones and applied both the baseline and ShapeMoiré methods to these cases, as depicted in Fig.~\ref{fig:vis-real}. It is important to note that the patterns of moiré are not fixed; for the same original image, different devices may capture different moiré patterns, as illustrated by the two images with moiré, where $I^A_{moir\acute{e}}$ is the original input image in the UHDM dataset~\cite{yu2022towards} and $I^B_{moir\acute{e}}$ is from our captured smartphone input image. It can be observed that $I^A_{moir\acute{e}}$ exhibits more widespread moiré artifacts across the entire region ($Patttern^A$), while in $I^B_{moir\acute{e}}$, the artifacts are concentrated in the sky area at the center of the image ($Patttern^B$). From the results in Fig.~\ref{fig:vis-real}, several observations can be made: 1) Our ShapeMoiré consistently outperforms the baseline method, regardless of the two morphologies of moiré, $I^A_{moir\acute{e}}$ and $I^B_{moir\acute{e}}$; our method exhibits superior moiré removal effects. 2) Our method demonstrates robustness against different moiré patterns, even helping alleviate color distortions. As shown in example $I^B_{moir\acute{e}}$ of Fig.~\ref{fig:vis-real}, due to severe moiré in the blank sky area at the center, noticeable color deviations occur. ShapeMoiré not only excels in removing Moiré patterns but also effectively restores the true colors of the sky.

\begin{figure*}[h!]
	\centering
	\centering
	\includegraphics[width=1.0\textwidth]{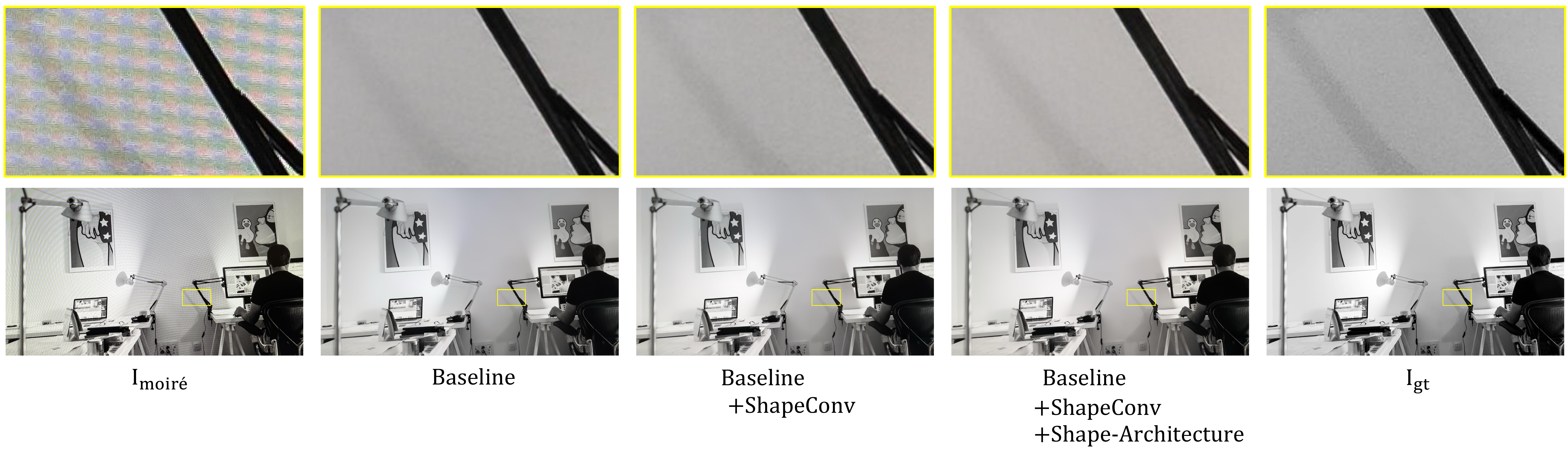}
    \vspace{-0.7em}
	\caption{\jm{Qualitative ablation results on the UHDM dataset using ESDNet as the baseline. From left to right: moiré-contaminated input, baseline result, baseline with ShapeConv, ShapeMoiré, and ground truth. The top row of each image shows a zoom-in of the highlighted region, allowing for closer inspection of visual differences.}
 }
	\label{fig:ablation_vis}
 \end{figure*}
 
 \jm{In Fig.~\ref{fig:ablation_vis}, we present qualitative comparisons to visually examine the effects of ablating key components, including ShapeConv and the Shape-Architecture, based on the strong ESDNet\cite{yu2022towards} baseline. As shown, while the baseline already delivers competitive results due to its advanced design, further improvements can still be observed when our proposed modules are added. Specifically, adding ShapeConv leads to better restoration of fine-grained details. For instance, in the shadow area beneath the lamp structure, ShapeConv enables more accurate reconstruction compared to the baseline alone. This suggests that the patch-level shape-aware representation contributes to finer spatial precision. Moreover, incorporating the Shape-Architecture introduces a noticeable enhancement in global image appearance. Compared to the baseline’s overall cooler tone, the Shape-Architecture helps correct color bias, producing a warmer tone that more closely aligns with the ground truth. This demonstrates its ability to recover higher-level structural and color features from the image-level shape stream. Together, these qualitative results visually confirm the effectiveness of each component, serving as a strong perceptual complement to our quantitative evaluations.}
\section{Conclusion}
\label{sec:con}

In this paper, we extend our previously accepted ShapeConv to the image demoiréing domain. 
Unlike existing methods, our proposed ShapeMoiré addresses the channel inconsistency problem of moiré patterns using \emph{shape} features. 
The key extension in the method is the introduction of the Shape-Architecture, which models the \emph{shape} information from the image-level and acts as a complement to the patch-level ShapeConv. 
The experimental results demonstrate its superiority over existing state-of-the-arts on four public popular datasets. 
Furthermore, we verify that our method can be painlessly integrated into four strong baselines to improve their performance, while introducing negligible computational cost during inference. 
Additionally, our ShapeMoiré yields significant promise in real-world cases with irregular moiré patterns, making it potentially suitable for edge device application and deployment.
\jm{Beyond demoiréing, we have also applied ShapeMoiré to the image deblurring task and observed promising improvements, suggesting its potential generalizability to broader image restoration problems.}

\jm{
Despite the encouraging results, our method has certain limitations. For instance, the current design primarily targets repetitive moiré patterns and may be less effective for more complex or irregular noise types that do not exhibit clear shape structures. Also, while computational overhead during inference is minimal, training with ShapeConv and Shape-Architecture requires additional resources.
In future work, we plan to explore adapting and extending the framework to other image restoration tasks such as general denoising, deraining, and compression artifact removal. Furthermore, integrating more advanced frequency-domain analysis and lightweight model designs could improve both effectiveness and efficiency, making the approach more practical for diverse real-world applications.}

\begin{acks}

This work is fully supported by the Advanced Research and Technology Innovation Centre (ARTIC), the National University of Singapore under Grant (project number: ELDT-RP2).
\end{acks}

\bibliographystyle{ACM-Reference-Format}
\bibliography{main}

\appendix
\section{Proof for Equation 9}
\label{sec:proof}
\begin{proof}\renewcommand{\qedsymbol}{}
\begin{equation*}
\scalebox{0.75}{
$
\begin{aligned}
    \mathbb{F}_{c_{out}} = &\sum_{k}^{K_h \times K_w \times C_{in}} (\mathbb{K}_{k,{c_{out}}} \times \textbf{P}_{\textbf{BS}_{k}})\\
    = &\sum_{i}^{K_h \times K_w} \sum_{j}^{C_{in}} (\mathbb{K}_{i,j,{c_{out}}} \times \textbf{P}_{\textbf{BS}_{i,j}})\\
    = &\sum_{i}^{K_h \times K_w} \sum_{j}^{C_{in}} (\mathbb{K}_{i,j,{c_{out}}} \times (\textbf{P}_{\textbf{B}_{1,j}} +\textbf{P}_{\textbf{S}_{i,j}}))\\
    = &\sum_{i}^{K_h \times K_w} \sum_{j}^{C_{in}} (\mathbb{K}_{i,j,{c_{out}}} \times (\mathbb{W}_B \times \mathbb{P}_{B_{1,j}} + \sum_{m}^{K_h \times K_w} (\mathbb{W}_{S_{m,i,j}} \times \mathbb{P}_{S_{m,j}}))\\
    = &\sum_{i}^{K_h \times K_w} \sum_{j}^{C_{in}} (\mathbb{W}_B \times \mathbb{K}_{i,j,{c_{out}}} \times \mathbb{P}_{B_{1,j}} + \sum_{m}^{K_h \times K_w} (\mathbb{W}_{S_{m,i,j}} \times \mathbb{K}_{i,j,{c_{out}}} \times \mathbb{P}_{S_{m,j}}))\\
    = &\sum_{i}^{K_h \times K_w} \sum_{j}^{C_{in}} (\mathbb{W}_B \times \mathbb{K}_{B_{1,j,{c_{out}}}} \times \mathbb{P}_{i,j} + \sum_{m}^{K_h \times K_w} (\mathbb{W}_{S_{m,i,j}} \times \mathbb{K}_{i,j,{c_{out}}} \times (\mathbb{P}_{m,j}-\mathbb{P}_{B_{1,j}}))\\
    = &\sum_{i}^{K_h \times K_w} \sum_{j}^{C_{in}} (\textbf{K}_{\textbf{B}_{1,j,{c_{out}}}} \times \mathbb{P}_{i,j}) + \sum_{i}^{K_h \times K_w} \sum_{j}^{C_{in}} \sum_{m}^{K_h \times K_w} (\mathbb{W}_{S_{m,i,j}} \times \mathbb{K}_{i,j,{c_{out}}} \times \mathbb{P}_{m,j} - \mathbb{W}_{S_{m,i,j}} \times \mathbb{K}_{i,j,{c_{out}}} \times \mathbb{P}_{B_{1,j}}))\\
    = &\sum_{i}^{K_h \times K_w} \sum_{j}^{C_{in}} (\textbf{K}_{\textbf{B}_{1,j,{c_{out}}}} \times \mathbb{P}_{i,j})+ \sum_{i}^{K_h \times K_w} \sum_{j}^{C_{in}} \sum_{m}^{K_h \times K_w} (\mathbb{W}_{S_{m,i,j}} \times \mathbb{K}_{i,j,{c_{out}}} \times \mathbb{P}_{m,j} - \mathbb{W}_{S_{m,i,j}} \times \mathbb{K}_{B_{1,j,{c_{out}}}} \times \mathbb{P}_{m,j}))\\
    = &\sum_{i}^{K_h \times K_w} \sum_{j}^{C_{in}} (\textbf{K}_{\textbf{B}_{1,j,{c_{out}}}} \times \mathbb{P}_{i,j}) + \sum_{i}^{K_h \times K_w} \sum_{j}^{C_{in}} \sum_{m}^{K_h \times K_w} (\mathbb{W}_{S_{m,i,j}} \times (\mathbb{K}_{i,j,{c_{out}}} - \mathbb{K}_{B_{1,j,{c_{out}}}}) \times \mathbb{P}_{m,j})\\
     =& \sum_{i}^{K_h \times K_w} \sum_{j}^{C_{in}} (\textbf{K}_{\textbf{B}_{1,j,{c_{out}}}} \times \mathbb{P}_{i,j}) + \sum_{i}^{K_h \times K_w} \sum_{j}^{C_{in}} \sum_{m}^{K_h \times K_w} (\mathbb{W}_{S_{m,i,j}} \times \mathbb{K}_{S_{i,j,{c_{out}}}} \times \mathbb{P}_{m,j})\\
     = &\sum_{i}^{K_h \times K_w} \sum_{j}^{C_{in}} (\textbf{K}_{\textbf{B}_{1,j,{c_{out}}}} \times \mathbb{P}_{i,j})+ \sum_{m}^{K_h \times K_w} \sum_{j}^{C_{in}} (\textbf{K}_{\textbf{S}_{m,j,{c_{out}}}} \times \mathbb{P}_{m,j})\\
     =&\sum_{i}^{K_h \times K_w} \sum_{j}^{C_{in}} (\textbf{K}_{\textbf{B}_{1,j,{c_{out}}}} \times \mathbb{P}_{i,j} + \textbf{K}_{\textbf{S}_{i,j,{c_{out}}}} \times \mathbb{P}_{i,j})\\
     =& \sum_{i}^{K_h \times K_w} \sum_{j}^{C_{in}} ((\textbf{K}_{\textbf{B}_{1,j,{c_{out}}}} + \textbf{K}_{\textbf{S}_{i,j,{c_{out}}}})\times \mathbb{P}_{i,j})\\
     =& \sum_{i}^{K_h \times K_w} \sum_{j}^{C_{in}} (\textbf{K}_{\textbf{BS}_{i,j,{c_{out}}}} \times \mathbb{P}_{i,j})\\
     =& \sum_{k}^{K_h \times K_w \times C_{in}} (\textbf{K}_{\textbf{BS}_{k,{c_{out}}}} \times \mathbb{P}_{k})
\end{aligned}
$
}
\end{equation*}
\end{proof}

\end{document}